\documentclass[runningheads]{llncs}
\usepackage{graphicx,amsmath,amsfonts,color,xcolor,CJK,multirow,adjustbox}
\usepackage{graphicx}
\usepackage{amsmath}
\usepackage{amsfonts}
\usepackage{color}
\usepackage{xcolor}
\usepackage{CJK}
\usepackage{multirow}
\usepackage{adjustbox}
\usepackage{marvosym}

\definecolor{citecolor}{RGB}{0, 113, 188}
\usepackage[pagebackref,breaklinks,colorlinks,citecolor=citecolor]{hyperref}

\usepackage{booktabs} 
\usepackage{colortbl} 
\usepackage{graphicx} 
\usepackage{subfigure} 
\usepackage{enumitem}  
\usepackage{xspace}  
\usepackage[bottom]{footmisc}   

\makeatletter
\DeclareRobustCommand\onedot{\futurelet\@let@token\@onedot}
\def\@onedot{\ifx\@let@token.\else.\null\fi\xspace}

\makeatother

\newcommand{\vprompt}[0]{\textsc{VPT}}

\newcommand{\partialft}[0]{\textsc{Partial}}
\newcommand{\linear}[0]{\textsc{Linear}}
\newcommand{\fullft}[0]{\textsc{Full}}
\newcommand{\sidetune}[0]{\textsc{Sidetune}}
\newcommand{\mlp}[0]{\textsc{Mlp}}
\newcommand{\bias}[0]{\textsc{Bias}}
\newcommand{\adapter}[0]{\textsc{Adapter}}

\definecolor{tabvline}{HTML}{a8a495}
\definecolor{prompt_blue}{HTML}{1f78b4}
\definecolor{prompt_red}{HTML}{d45c43}

\definecolor{green_im}{rgb}{0.0, 0.5, 0.0}
\newcommand{\ttbf}[1]{\textbf{\texttt{#1}}}
\newcommand{\band}{\rowcolor{gray!15}}

\definecolor{citecolor}{RGB}{0, 113, 188}

\usepackage[capitalize]{cleveref}
\crefname{section}{Sec.}{Secs.}
\Crefname{section}{Section}{Sections}
\Crefname{table}{Table}{Tables}
\crefname{table}{Tab.}{Tabs.}

\makeatletter
\newcommand{\printfnsymbol}[1]{%
  \textsuperscript{\@fnsymbol{#1}}%
}
\makeatother

\begin{document}
\title{Dynamic Visual Prompt Tuning for Parameter Efficient Transfer Learning}
\titlerunning{DVPT}
%

\author{Chunqing Ruan\inst{1} \and
Hongjian Wang\inst{2}\textsuperscript{(\Letter)}}
\authorrunning{C. Ruan et al.}
%
\institute{Beijing University of Posts and Telecommunications, Beijing, China \\
\email{chunqing020200@gmail.com}  \and
Jilin University, Jilin, China\\
\email{forever020200@gmail.com}}

%
%

%
\maketitle              
\begin{abstract}
Parameter efficient transfer learning (PETL) is an emerging research spot that aims to adapt large-scale pre-trained models to downstream tasks. Recent advances have achieved great success in saving storage and computation costs. However, these methods do not take into account instance-specific visual clues for visual tasks. In this paper, we propose a Dynamic Visual Prompt Tuning framework (DVPT), which can generate a dynamic instance-wise token for each image. In this way, it can capture the unique visual feature of each image, which can be more suitable for downstream visual tasks. We designed a Meta-Net module that can generate learnable prompts based on each image, thereby capturing dynamic instance-wise visual features. Extensive experiments on a wide range of downstream recognition tasks show that DVPT achieves superior performance than other PETL methods. More importantly, DVPT even outperforms full fine-tuning on 17 out of 19 downstream tasks while maintaining high parameter efficiency. Our code will be released soon.

\keywords{parameter efficient transfer learning (PETL) \and instance-specific visual clues \and Meta-Net \and Dynamic Visual Prompt Tuning (DVPT).}
\end{abstract}
\section{Introduction}
Recently, the Transformer~\cite{ref_paper1} has shown significant potential in achieving various objectives, including natural language processing(NLP)~\cite{ref_paper2}, visual recognition~\cite{ref_paper6,ref_paper7}, dense prediction~\cite{ref_paper10}, Generative Adversarial Network (GAN)~\cite{ref_paper15,ref_paper16}, reinforcement learning~\cite{ref_paper17}, robotics~\cite{ref_paper20}, etc. Especially, the large models based on transformer exhibit extremely strong generalization performance. Currently, there is a growing interest in adapting large pre-trained models to various downstream tasks, since this approach offers the advantage of reducing the need for designing and training task-specific models.

Existing literature related to adapting pre-trained model~\cite{ref_paper22,ref_paper23} in computer vision tend to focus on two prevalent approaches, fine-tuning and linear probing. Fine-tuning involves updating all the model parameters based on new datasets, which typically leads to better performance than linear probing on various tasks~\cite{ref_paper24}. However, fine-tuning can be computationally expensive and parameter inefficient, especially for larger models like Transformers, whose parameters grow exponentially. On the other hand, linear probing only updates and stores new prediction heads while keeping the backbone frozen, making it more computational and parameter efficient. However, linear probing often results in inferior performance compared to fine-tuning.

Recently, there have been some new techniques to overcome such a dilemma, such as adapter~\cite{ref_paper28,ref_paper29}, LoRA~\cite{ref_paper33}, and NOAH~\cite{ref_noah}. These PETL methods add lightweight modules to pre-trained models, freeze the pre-trained weights, and fine-tune the model to adapt to downstream tasks. Another approach is based on Visual Prompt Tuning (VPT)~\cite{ref_vpt}. It introduces a small number of learnable prompts into the embeddings of image patches while freezing the whole backbone during downstream training. However, the learnable prompts are fixed and do not consider instance-specific visual clues. It is updated entirely through the network's training process, so it is difficult for the network to capture visual features. Therefore, a well-designed vision-oriented PETL method is expected to introduce additional inductive bias and visual instance-wise features.

To overcome the issue, we proposed a dynamic visual prompt tuning framework (DVPT), which can adaptively learn a unique prompt for each image instead of a fixed prompt. To ensure that the model remains parameter efficient, we implement DVPT through a lightweight Meta-Net module, which generates learnable prompts for each image. The learnable prompts are designed to accurately represent each image, rather than only serving certain classes. Therefore, DVPT can fully leverage each input image's visual features, making it more suitable for downstream visual tasks. We validate DVPT on 19 downstream recognition tasks using a pre-trained ViT as the backbone, and the results show that DVPT achieves higher accuracy than other PETL methods. 

Overall, we summarize our contributions as follows. (1) We point out that existing PETL methods ignore the visual instance-wise feature, which limits their transferability across different tasks. (2) We propose DVPT, a simple yet effective PETL method that generates dynamic prompts using instance-specific visual clues to adapt pre-trained models to downstream visual tasks. (3) Extensive results on VTAB show that DVPT outperforms full fine-tuning and previous PETL methods, validating the effectiveness of DVPT.

\section{Related Works}
The DVPT approach incorporates a plug-and-play Meta-Net module to fine-tune the current vision Transformer models. To provide context for this approach, we perform a literature review focused on two key areas: the vision Transformers and efficient transfer learning for vision Transformers.

\subsection{Transformer in Vision}
The Transformer architecture was introduced in~\cite{ref_paper1} and then revolutionized the field of natural language processing~\cite{ref_paper2}. The remarkable achievements of Transformers in natural language processing have also inspired computer vision researchers to adopt this architecture since Vision Transformer (ViTs)~\cite{ref_paper6}. The Transformer's ability to effectively model long-range relationships has proven useful in a variety of computer vision tasks, including image classification~\cite{ref_paper6,ref_paper7}, object detection~\cite{ref_paper10,ref_object_detection1}, semantic/instance segmentation~\cite{ref_paper12}, video understanding~\cite{ref_video_understanding1,ref_video_understanding2}, point cloud modeling~\cite{ref_point_cloud1,ref_point_cloud2}, 3D object recognition~\cite{ref_3d} and even low-level image processing~\cite{ref_low_level1,ref_low_level2}. Transformer models have further improved performance in vision recognition by large-scale pretraining~\cite{ref_pretraining1,ref_pretraining2}. With the advent of pre-trained Transformer models significantly larger than previous CNN backbones, a key challenge is how to fine-tune these models for downstream vision tasks effectively.

\subsection{Efficient Transfer learning for Transformers}
Researchers are increasingly interested in PETL to optimize large-scale pre-trained models. PETL was first used in natural language processing~\cite{ref_paper28,ref_paper33}, showing that fine-tuning lightweight modules in a large pre-trained model can achieve almost optimal performance. Based on this success, PETL principles have been applied to large pre-trained vision models for various vision tasks ~\cite{ref_adaptformer,ref_paper28,ref_noah,ref_paper32}. Two main approaches include adapter-based methods~\cite{ref_adaptformer,ref_paper28}, such as Adaptformer, and prompt tuning-based methods~\cite{ref_cocoop,ref_paper32}, such as VPT. Adapter-based methods insert small MLP networks into the vision model to adapt to downstream tasks, while prompt tuning-based methods add trainable tokens to the input sequence of the vision Transformer to reduce the gap between pre-training and downstream data distributions. LoRA~\cite{ref_paper33} obtains a low-rank representation of multi-head attention~\cite{ref_paper1} while keeping the parameters frozen. Zhang~\cite{ref_noah} et al. propose a prompt search algorithm that combines adapter, prompt tuning, and LoRA automatically.

\begin{figure}
\includegraphics[width=\textwidth]{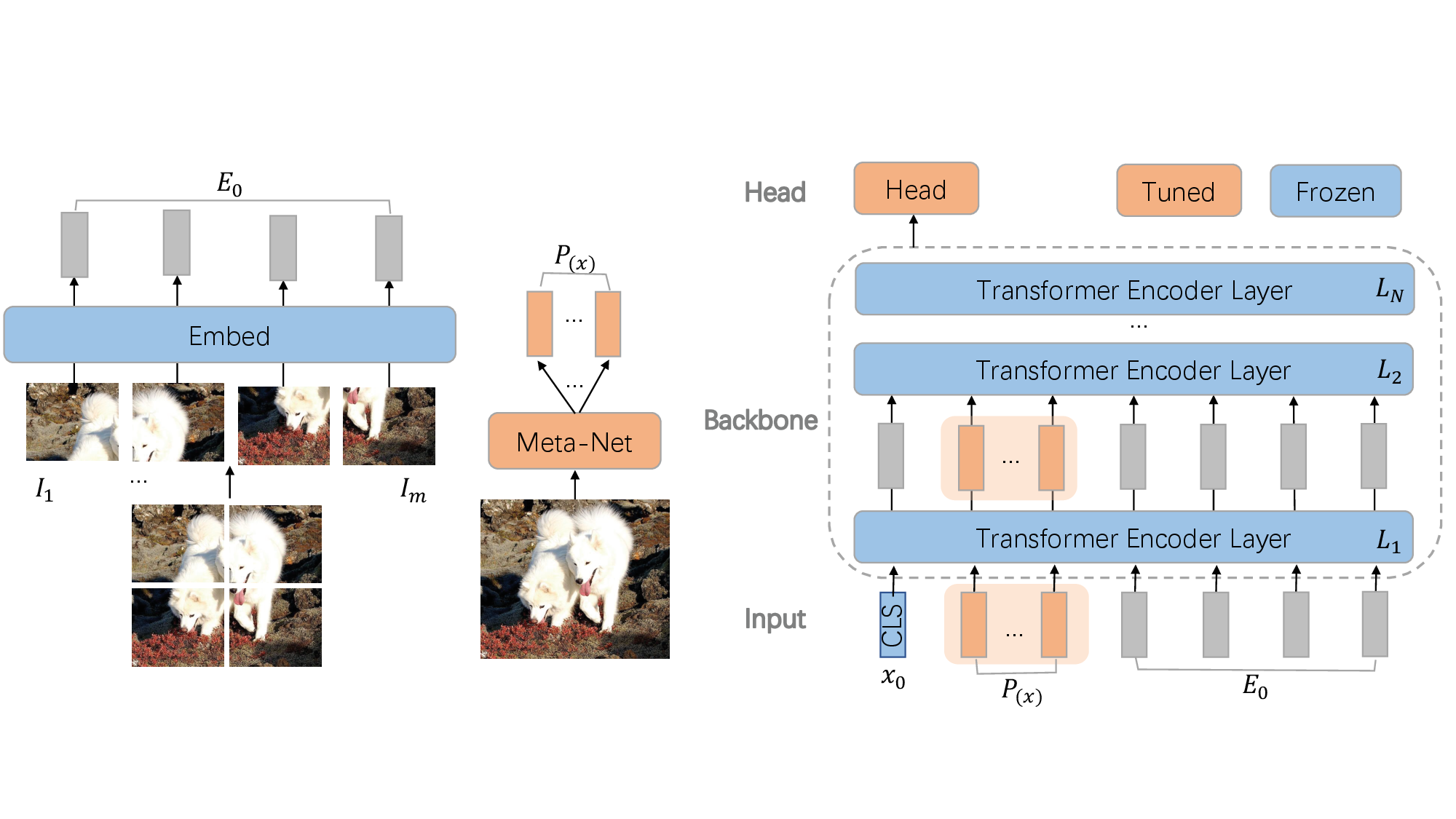}
\caption{Our approach, Dynamic Visual Prompt Tuning (DVPT), consists of three learnable components: a set of prompts, linear head and a lightweight Meta-Net that generates dynamic prompts for each input image. During training on downstream tasks, only the learnable components are updated while the whole Transformer encoder is frozen.} \label{fig}
\end{figure}

\section{Approach}
We propose the DVPT for adapting large, pre-trained vision transformer models to downstream tasks. The overall framework is presented in Fig.~\ref{fig}.

\subsection{Preliminaries}
\label{subsec:ViT}
\subsubsection{Vision Transformer.}
In a Vision Transformer (ViT) ~\cite{ref_paper6} with N layers, the input image $\mathbf{x}$ is partitioned into fixed-sized patches $\{I_j \in \mathbb{R}^{3 \times h \times w}|j \in \mathbb{N}, 1 \le j \le m\}$, where $h, w$ are the height and width of image patches and $m$ is the number of image patches. Each patch is embedded into a $d$-dimensional latent space and undergoes positional encoding:$$\mathbf{e}_0^j=Embed(I_j) \qquad \mathbf{e}_0^j \in \mathbb{R}^d,j=1,2,...m. \eqno(1)$$

We use image patch embeddings, $\mathbf{E}_i=\{ \mathbf{e}_i^j \in \mathbb{R}^d|j\in \mathbb{N}, 1 \le j \le m\}$ as inputs for the ${i+1}$-th Transformer layer $(L_{i+1})$. Together with an additional learnable classification token [CLS], the entire ViT can be expressed as follows,
\begin{align}
[\mathbf{x}_i,\mathbf{E}_i]&=L_i([\mathbf{x}_{i-1}, \mathbf{E}_0]) \qquad i=1,2,...,N \tag{2}\\
\mathbf{y}&=Head(\mathbf{x}_N), \tag{3}
\end{align}
where $\mathbf{x}_i \in \mathbb{R}^d$ is the embedding of [CLS] at $L_{i+1}$'s input space. The notation $[\cdot, \cdot]$ denotes stacking and concatenating along the dimension of sequence length, $i.e.,[\mathbf{x}_i, \mathbf{E}_i] \in \mathbb{R}^{(1+m) \times d}$. Each Transformer encoder has two sub-layers: a multi-head self-attention layer (MHSA) and an MLP layer. A neural classification head translates the [CLS] embedding, $\mathbf{x}_N$, from the final layer into a distribution of predicted class probabilities represented by $\mathbf{y}$. Additional information on this topic can be found in~\cite{ref_paper6}.

\subsubsection{Visual Prompt Tuning(VPT).}
 Given a pre-trained Transformer model, VPT adds a set of $p$ continuous embeddings of dimension $d$, i.e., \textit{prompts}, to the input space after the embedding layer. Only the task-specific prompts are updated during fine-tuning while the Transformer backbone remains frozen.

 Each prompt token can be represented as a learnable vector in a $d$-dimensional space. $\mathbf{P} = \{ \mathbf{p}^k \in \mathbb{R}^d|k \in \mathbb{N}, 1 \le k \le p\}$ denotes a collection of $p$ prompts. The prompted ViT is:
\begin{align}
[\mathbf{x}_1,\mathbf{Z}_1,\mathbf{E}_1]&=L_1([\mathbf{x}_0,\mathbf{P},\mathbf{E}_0]) \tag{4}\\
[\mathbf{x}_i,\mathbf{Z}_i,\mathbf{E}_i]&=L_i([\mathbf{x}_{i-1},\mathbf{Z}_{i-1},\mathbf{E}_{i-1}]) \qquad i=2,3,...,N \tag{5}\\
\mathbf{y}&=Head(\mathbf{x}_N), \tag{6}
\end{align}
where $\mathbf{Z}_i \in \mathbb{R}^{p \times d}$ refers to the features computed by the $i$-th Transformer layer, and $[\mathbf{x}_i,\mathbf{Z}_i,\mathbf{E}_i]\in\mathbb{R}^{(1+p+m)\times d}$.

\subsection{Dynamic Visual Prompt Tuning}
\label{subsec:DVPT}
As discussed in \cref{subsec:ViT}, VPT uses a fixed prompt for different images. We argue that visual instance-wise feature can achieve better performance because it is centered on analyzing each input, rather than relying on a fixed set of classes. Therefore, we propose an efficient approach DVPT. It extends VPT by further introducing a Meta-Net module, which generates instance-wise visual features for each input.

The input image $\mathbf{x}$ is fed into the Meta-Net module to get dynamic prompts. Inspired by the VPT~\cite{ref_vpt}, the dynamic prompts are the combination of learnable prompts and dynamic instance-wise visual clues, $\mathbf{P}(\mathbf{x})=\mathbf{P}+\boldsymbol{\pi}$. The $\mathbf{P} = \{ \mathbf{p}^k \in \mathbb{R}^d|k \in \mathbb{N}, 1 \le k \le p\}$ is a collection of $p$ prompts following VPT. Finally, the DVPT is formulated as:
\begin{align}
[\mathbf{x}_1,\mathbf{Z}_1,\mathbf{E}_1]&=L_1([\mathbf{x}_0,\mathbf{\color{red}P(\mathbf{x})},\mathbf{E}_0]) \tag{7}\\
[\mathbf{x}_i,\mathbf{Z}_i,\mathbf{E}_i]&=L_i([\mathbf{x}_{i-1},\mathbf{P}_{i-1},\mathbf{E}_{i-1}]) \qquad i=2,3,...,N \tag{8}\\
\mathbf{y}&=Head(\mathbf{x}_N), \tag{9}
\end{align}

 During training, we update both the Meta-Net parameters and the prompt vectors $\mathbf{P}(\mathbf{x})$. In our experiments, 4 linear layers are used to implement the Meta-Net module for efficiency. More complex networks can lead to better performance. The detailed analysis can be found in \cref{subsec:ablate}.

\section{Experiments}
We utilized pre-trained Transformer backbones to evaluate the effectiveness of DVPT in various downstream recognition tasks. We provide a detailed account of our experimental setup in \cref{subsec:evalsetup}, covering the pre-trained backbone, the downstream tasks, and an overview of alternative transfer learning methods. In \cref{subsec:exp_results}, we demonstrate our approach's effectiveness and practical utility. Furthermore, we conducted a thorough analysis of how different design choices can influence performance in \cref{subsec:ablate}, contributing to a better understanding of our method.

\begin{table*}[t]
\normalsize
\caption{ViT-B/16 pre-trained on supervised ImageNet-21k. For each method and each downstream task, we report the test accuracy score and the number of wins in $(\cdot)$ compared to Full. The best results among all methods except FULL are \textbf{bolded}.}\label{tab1}
\resizebox{\textwidth}{!}{
\begin{tabular}{
l|
ccccccc!{\color{tabvline}\vrule} c|
cccc!{\color{tabvline}\vrule} c|
cccccccc!{\color{tabvline}\vrule} c
}
\toprule
  &\rotatebox{90}{\bf{CIFAR-100}}
  &\rotatebox{90}{\bf{Caltech101} }
  &\rotatebox{90}{\bf{DTD} }
  &\rotatebox{90}{\bf{Flowers102} }
  &\rotatebox{90}{\bf{Pets} }
  &\rotatebox{90}{\bf{SVHN} }
  &\rotatebox{90}{\bf{Sun397} }
  &\rotatebox{90}{\bf{Mean}}
  &\rotatebox{90}{\bf{Patch Camelyon} }
  &\rotatebox{90}{\bf{EuroSAT} }
  &\rotatebox{90}{\bf{Resisc45} }
  &\rotatebox{90}{\bf{Retinopathy} }
  &\rotatebox{90}{\bf{Mean}}
  &\rotatebox{90}{\bf{Clevr/count} }
  &\rotatebox{90}{\bf{Clevr/distance} }
  &\rotatebox{90}{\bf{DMLab}}
  &\rotatebox{90}{\bf{KITTI/distance} }
  &\rotatebox{90}{\bf{dSprites/location} }
  &\rotatebox{90}{\bf{dSprites/orientation} }
  &\rotatebox{90}{\bf{SmallNORB/azimuth} }
  &\rotatebox{90}{\bf{SmallNORB/elevation} }
  &\rotatebox{90}{\bf{Mean}}
  \\
\midrule
\band \fullft{} &68.9 &87.7 &64.3 &97.2 &86.9 &87.4 &38.8 &75.88 &79.7 &95.7 &84.2 &73.9 &83.36 &56.3 &58.6 &41.7 &65.5 &57.5 &46.7 &25.7 &29.1 &47.64 
\\
\midrule
\linear{} &63.4 &85.0 &63.2 &97.0 &86.3 &36.6 &51.0 &68.93 (1) &78.5 &87.5 &68.6 &74.0 &77.16 (1) &34.3 &30.6 &33.2 &55.4 &12.5 &20.0 &9.6 &19.2 &26.84 (0)\\
\partialft{}-1 &66.8 &85.9 &62.5 &97.3 &85.5 &37.6 &50.6 &69.44 (2) &78.6 &89.8 &72.5 &73.3 &78.53 (0) &41.5 &34.3 &33.9 &61.0 &31.3 &32.8 &16.3 &22.4 &34.17 (0)
\\
\mlp{}-2 &63.2 &84.8 &60.5 &97.6 &85.9 &34.1 &47.8 &67.70 (2) &74.3 &88.8 &67.1 &73.2 &75.86 (0) &45.2 &31.6 &31.8 &55.7 &30.9 &24.6 &16.6 &23.3 &32.47 (0)
\\
\mlp{}-5 &59.3 &84.4 &59.9 &96.1 &84.4 &30.9 &46.8 &65.98 (1) &73.7 &87.2 &64.8 &71.5 &74.31 (0) &50.8 &32.3 &31.5 &56.4 &7.5 &20.8 &14.4 &20.4 &29.23 (0)
\\
\midrule
\sidetune{} &60.7 &60.8 &53.6 &95.5 &66.7 &34.9 &35.3 &58.21 (0) &58.5 &87.7 &65.2 &61.0 &68.12 (0) &27.6 &22.6 &31.3 &51.7 &8.2 &14.4 &9.8 &21.8 &23.41 (0)
\\
\bias{} &72.8 &87.0 &59.2 &97.5 &85.3 &59.9 &51.4 &73.30 (3) &78.7 &91.6 &72.9 &69.8 &78.25 (0) &61.5 &55.6 &32.4 &55.9 &66.6 &40.0 &15.7 &25.1 &44.09 (2)
\\
\adapter{}-64 &74.2 &85.8 &62.7 &97.6 &87.2 &36.3 &50.9 &70.65 (4) &76.3 &87.5 &73.7 &70.9 &77.10 (0) &42.9 &39.9 &30.4 &54.5 &31.9 &25.6 &13.5 &21.4 &32.51 (0)
\\
\adapter{}-8 &74.2 &85.7 &62.7 &97.8 &87.2 &36.4 &50.7 &70.67 (4) &76.9 &89.2 &73.5 &71.6 &77.80 (0) &45.2 &41.8 &31.1 &56.4 &30.4 &24.6 &13.2 &22.0 &33.09 (0)
\\
VPT &78.8 &90.8 &65.8 &\textbf{98.0} &88.3 &78.1 &49.6 &78.48 (6) &81.8 &96.1 &83.4 &68.4 &82.43 (2) &68.5 &60.0 &46.5 &72.8 &73.6 &47.9 &32.9 &37.8 &54.98 (8) 
\\
\midrule
DVPT &\textbf{80.3} &\textbf{91.9} &\textbf{67.6} &97.8 &\textbf{89.5} &\textbf{80.3} &\textbf{51.9} &\textbf{79.9(6)} &\textbf{83.8} &\textbf{96.9} &\textbf{85.3} &\textbf{70.5} &\textbf{84.13(3)} &\textbf{70.6} &\textbf{63.9} &\textbf{48.9} &\textbf{76.8} &\textbf{76.5} &\textbf{50.3} &\textbf{35.6} &\textbf{39.9} &\textbf{57.81(8)} 
\\
\bottomrule
\end{tabular}
}
\end{table*}

\subsection{Experiment Setup}
\label{subsec:evalsetup}

\subsubsection{Pre-trained Backbones.}
In the experiment, we utilize the Vision Transformer(ViT)~\cite{ref_paper6} as the Transformer architecture in vision. The backbone in this section is pre-trained on ImageNet-21k~\cite{ref_imagenet}. We have adhered to the original configurations, such as the number of image patches and the existence of [CLS] token.

\subsubsection{Baselines.}
We are evaluating how DVPT performs in comparison to other fine-tuning methods that are commonly used:

\begin{enumerate}[nosep, label=(\alph*), font=\small\ttbf,] 
\item \fullft{}: update all the parameters of both the backbone and classification head.

\item Methods concentrate on fine-tuning only the classification head while keeping the weights of the pre-trained backbone frozen.
\begin{itemize}[leftmargin=0.0em, topsep=0.15mm]
\item \linear{}: utilize a single linear layer as the trainable classification head.
\item \partialft{}-$k$: fine-tune the last $k$ layers of backbone while freezing the others. It sets a new boundary between the backbone and the classification head.
\item \mlp{}-$k$: use a multilayer perceptron (MLP) with $k$ layers as the classification head instead of a linear layer.
\end{itemize}

\item Methods that add new learnable parameters to the backbone during fine-tuning process:
\begin{itemize}[leftmargin=0.0em, topsep=0.15mm]
\item \sidetune{}~\cite{ref_sidetune}: develop a "side" network and perform linear interpolation between pre-trained features and side-tuned features before being fed into the head.
\item \bias{}~\cite{ref_bias}: fine-tune only the bias terms of a pre-trained backbone.
\item \adapter{}~\cite{ref_adaptformer,ref_paper28}: introduce new MLP modules with the residual connection into Transformer layers.
\item \vprompt{}~\cite{ref_vpt}: add a set of prompts in the input space after the Embedding layer.
\end{itemize}
\end{enumerate}

\subsubsection{Downstream Tasks.}
The VTAB-1K benchmark consists of 19 image classification tasks from various domains, which can be broadly classified into three groups: Natural, Specialized, and Structured. Each task has only 1,000 training samples. We perform an 80/20 split on the 1000 training images in each task for hyperparameters searching. We run the final evaluation using the entire training data. The average accuracy score on the test set is reported based on three runs.

FGVC includes 5 established Fine-Grained Visual Classification tasks, namely CUB-200-2011, NABirds, Oxford Flowers, Stanford Dogs, and Stanford Cars. If a certain dataset only provides train and test sets, we randomly split the training set into 90\% for training and 10\% for validation. We use the validation set to select the best hyperparameters for our models.

\subsection{Main Results}
\label{subsec:exp_results}
\cref{tab1} displays the results of fine-tuning a pre-trained ViT-B/16 across three different downstream task groups. Our DVPT is compared with 8 other tuning methods. We can see that:

\begin{enumerate}[nosep, leftmargin=5mm]
\item DVPT achieves better results than \fullft{} (as shown in \cref{tab1}) on 17 out of 19 tasks, while using a significantly small number of total model parameters. The underline reason is the \fullft{} updates all parameters, destroying the knowledge learned by the pre-trained model, while DVPT does not change the weights of the pre-trained model and only learns some tokens, improving performance based on the pre-trained model. Thus, the DVPT approach shows great potential for adapting larger Transformers in vision tasks.

\item As shown in \cref{tab1}, DVPT surpasses all other parameter-efficient tuning methods across all task groups. Our DVPT outperforms the current SOTA PETL method, VPT, on 18 out of 19 VTAB-1K datasets, specifically, DTD(3.1\%), Retinopathy(2.12\%), and Clevr/distance(3.86\%). It validates the effectiveness of our DVPT when working with limited storage resources. Moreover, we discover that instance-wise visual clues play a key role in visual tasks.
\end{enumerate}

\setlength{\tabcolsep}{5pt}
\begin{table}[t]
\normalsize
\begin{center}
\caption{Ablation on the number of linear layers in the Meta-Net. We vary the number of linear layers in the DVPT and show the top-1 accuracy on 5 VTAB datasets. The best results among all numbers of linear layers are \textbf{bolded}.}\label{tab2}
\resizebox{0.6\textwidth}{!}{
\begin{tabular}{
c !{\color{tabvline}\vrule}
ccccc }
\toprule
Number of layers &DTD &EuroSAT &KITTI/distance &Pets &Resisc45 \\
\midrule
2 & 63.4 & 92.59 & 75.95 & 87.05 & 83.92 \\
4 & 67.61 & 96.91 & 76.79 & 89.52 & 85.29 \\
6 & \textbf{68.73} & \textbf{97.02} & 7\textbf{8.62} & \textbf{91.24} & \textbf{87.05} \\
\bottomrule
\end{tabular}
}
\end{center}
\end{table}
\setlength{\tabcolsep}{1.4pt}

\subsection{Ablation on Model Design Variants}
\label{subsec:ablate}
We ablate our DVPT to study what properties make for a good DVPT. The ablation studies conducted in this work are all performed on the VTAB-1K, with the same setup in \cref{tab1}.

\begin{figure}[t]
\centering
\includegraphics[width=\textwidth]{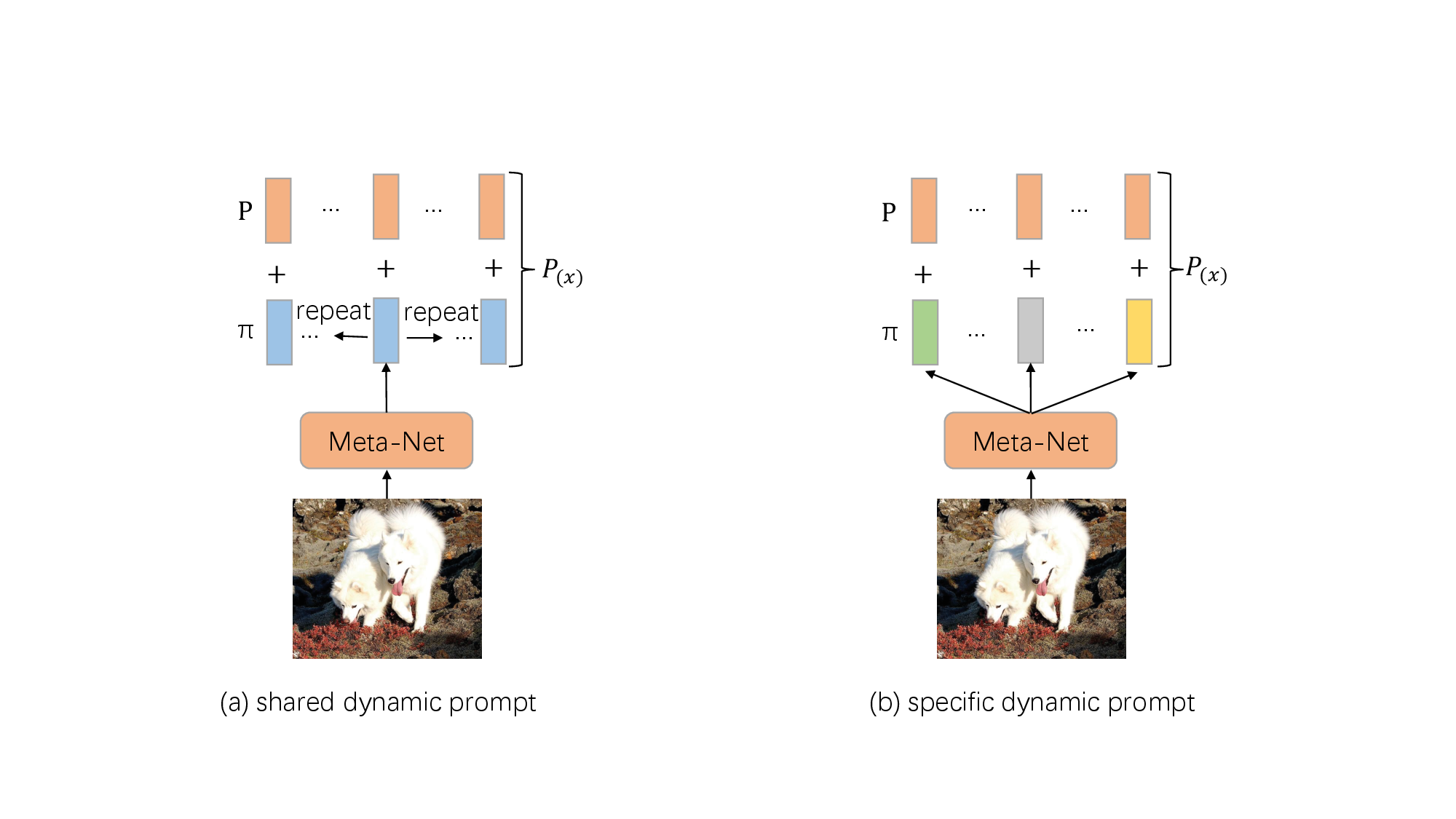}
\caption{
Two implementation approaches for dynamic prompts: (a) shared dynamic prompt (b) specific dynamic prompt.
}
\label{fig2}
\end{figure}

\subsubsection{Different number of linear layers in the Meta-Net module.}
The linear layers number of a Meta-Net controls the number of introduced parameters in DVPT. Using fewer linear layers can lead to a lower parameter count, with a possible performance cost. We ablate DVPT on the linear layers number to study these effects. Task results on different linear layers number are presented in \cref{tab2}. The result shows a consistent improvement in top-1 accuracy as the number of linear layers increases, e.g., for Resisc45, DVPT with 6 linear layers surpasses 4 linear layers and 2 linear layers by 1.76\%, 3.13\% on top-1 accuracy, respectively. In other experiments, we set the number of layers to 4 to achieve the trade-off between performance and parameter count. 

\subsubsection{Varying the dimension of instance-wise visual clues.}
As mentioned in section \cref{subsec:DVPT}, the output of the Meta-Net module is a \textit{shared dynamic prompt} (\cref{fig2} (a)). In this section, we let the Meta-Net module outputs a \textit{specific dynamic prompt}, as in \cref{fig2} (b). We study which case can achieve better performance. As shown in \cref{tab3}, we observe that the specific dynamic prompt method outperforms the shared dynamic prompt method, e.g., the specific method achieves 2.74\% top-1 accuracy gains over the shared method on KITTI/distance dataset.

\setlength{\tabcolsep}{5pt}
\begin{table}[t]
\begin{center}
\caption{Ablation on the dimension of instance-wise visual clues in the Meta-Net. We show the top-1 accuracy on 3 VTAB datasets. The best results are \textbf{bolded}.}\label{tab3}
\resizebox{0.7\textwidth}{!}{
\begin{tabular}{
c !{\color{tabvline}\vrule}
ccc }
\toprule
Method &KITTI/distance &Pets &Resisc45 \\
\midrule
shared dynamic prompt  &76.79 &89.52 &85.29 \\
specific dynamic prompt &\textbf{79.53} &\textbf{91.05} &\textbf{87.37} \\
\bottomrule
\end{tabular}
}
\end{center}
\end{table}
\setlength{\tabcolsep}{1.4pt}

\subsubsection{Different downstream data sizes.}
\begin{figure}[h]
\centering
\includegraphics[width=0.6\textwidth]{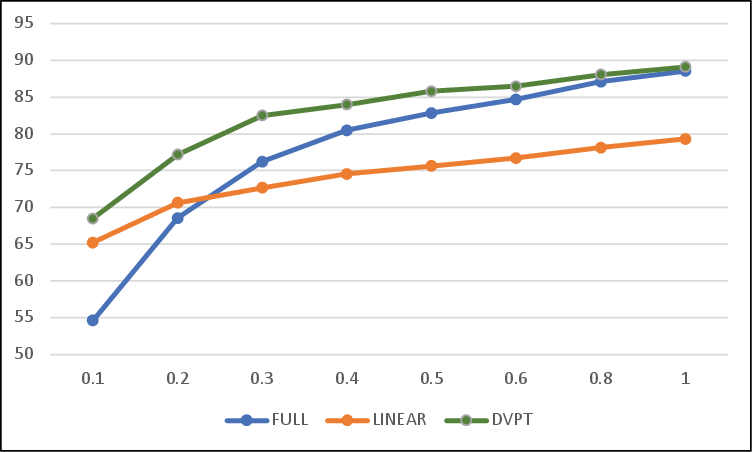}
\caption{
Performance comparison on different downstream data scales averaged across 5 FGVC tasks. DVPT is compared with Full and Linear.
}
\label{fig4}
\end{figure}

We further study the performance of DVPT under varying training data sizes on the FGVC dataset. Following VPT, we reduce the training data sizes to {10\%, 20\%, 30\%, 40\%, 50\%, 60\%, and 70\%} and compare DVPT with the Full and Linear methods. The results for each method on different training data scales are presented in~\cref{fig4}. It shows that DVPT consistently yields better performance than the Full and Linear across data scales. When there is limited data available, DVPT and Linear perform better than Full. However, as more training data becomes available, Full begins to outperform Linear. In contrast, DVPT still consistently outperforms Full across training data sizes.

\section{Conclusions}
In this paper, we point out that the current PETL methods ignore the visual instance-wise feature, which leads to suboptimal performance on downstream tasks. Based on the above findings, we propose a DVPT framework, a vision-oriented PETL method that employs Meta-Net to adapt pre-trained ViT to downstream tasks. In this way, DVPT can capture the unique visual features of each image, which can be more suitable for downstream visual tasks. The experimental results on the VTAB-1K benchmark indicate that DVPT achieves better performance than other PETL methods. Our approach highlights the importance of the visual instance-wise feature for visual tasks during the development of PETL techniques. Our work will encourage further research into developing more effective methods for fine-tuning large vision models.

\end{document}